  \documentclass[10pt]{article}
  \usepackage{amsmath,amsthm,amsfonts,amssymb,latexsym,graphicx,yhmath}
  \usepackage[T2A]{fontenc}
  \usepackage[utf8]{inputenc}
  \usepackage{algorithm}
  \usepackage{algpseudocode}
  \usepackage[pdfpagemode=UseNone,pdfstartview=FitH]{hyperref}

\allowdisplaybreaks[4]

  \newtheorem{lemma}{Lemma}

  \theoremstyle{definition}

  \newcommand{\OCMI}{Vovk/etal:2009AOS}

  \newcommand{\OCMVII}{Vovk/Petej:arXiv1211}

  \newcommand{\OCMX}{Burnaev/Vovk:arXiv1404}

  \newcommand{\OCMXVII}{Vovk/etal:2017COPA}
  \newcommand{\OCMXVIII}{Vovk:arXiv1708}

\newcommand{\R}{\mathbb{R}}

\DeclareMathOperator{\Expect}{\mathbb{E}}

\DeclareMathOperator{\cov}{cov}
\DeclareMathOperator{\corr}{corr}

\newcommand{\dd}{\mathrm{d}}

\newcommand{\hatbar}[1]{\widetriangle{#1}}

  \title{Conformal predictive distributions with kernels}
  \author{Vladimir Vovk, Ilia Nouretdinov,\\Valery Manokhin, and Alex Gammerman}

\begin{document}
\maketitle

\begin{abstract}
  This paper reviews the checkered history of predictive distributions in statistics
  and discusses two developments, one from recent literature and the other new.
  The first development is bringing predictive distributions into machine learning,
  whose early development was so deeply influenced by two remarkable groups at the Institute of Automation and Remote Control.
  The second development is combining predictive distributions with kernel methods,
  which were originated by one of those groups, including Emmanuel Braverman.

    \bigskip

    \noindent
    This paper has been prepared for the Proceedings of the Braverman Readings,
    held in Boston on 28--30 April 2017.
    Its version at \href{http://alrw.net}{http://alrw.net}
    (Working Paper 20, first posted on 19 October 2017) is updated most often.
\end{abstract}

\section{Introduction}

Prediction is a fundamental and difficult scientific problem.
We limit the scope of our discussion by imposing, from the outset,
two restrictions:
we only want to predict one real number $y\in\R$,
and we want  our prediction to satisfy a reasonable property of validity
(under a natural assumption).
It can be argued that the fullest prediction for $y$ is a probability measure on $\R$,
which can be represented by its distribution function:
see, e.g., \cite{Dawid:1984,Dawid/Vovk:1999,Gneiting/Katzfuss:2014}.
We will refer to it as the predictive distribution.
A standard property of validity for predictive distributions is being well-calibrated.
Calibration can be defined as the
``statistical compatibility between the probabilistic forecasts and the realizations''
\cite[Section~1.2]{Gneiting/Katzfuss:2014},
and its rough interpretation is that predictive distributions should tell the truth.
Of course, truth can be uninteresting and non-informative,
and there is a further requirement of efficiency,
which is often referred to as sharpness \cite[Section~2.3]{Gneiting/Katzfuss:2014}.
Our goal is to optimize the efficiency subject to validity
\cite[Section~1.2]{Gneiting/Katzfuss:2014}.

This paper is a very selective review of predictive distributions with validity guarantees.
After introducing our notation and setting the prediction problem
in Section~\ref{sec:setting},
we start, in Section~\ref{sec:Bayesian}, from the oldest approach to predictive distributions, Bayesian.
This approach gives a perfect solution but under a very restrictive assumption:
we need a full knowledge of the stochastic mechanism generating the data.
In Section~\ref{sec:fiducial} we move to Fisher's fiducial predictive distributions.

The first recent development (in~\cite{\OCMXVII}, as described in Section~\ref{sec:conformal} of this paper)
was to carry over predictive distributions to the framework of statistical machine learning as developed by two groups
at the Institute of Automation and Remote Control
(Aizerman's laboratory including Braverman and Rozonoer and Lerner's laboratory including Vapnik and Chervonenkis;
for a brief history of the Institute and research on statistical learning there,
including the role of Emmanuel Markovich Braverman,
see \cite{Vovk/etal:2015Chervonenkis}, especially Chapter~5).
That development consisted in adapting predictive distributions to the IID model,
discussed in detail in the next section.
The simplest linear case was considered in \cite{\OCMXVII}, with groundwork laid in \cite{\OCMX}.
The second development, which is this paper's contribution, is combination with kernel methods,
developed by the members of Aizerman's laboratory,
first of all Braverman and Rozonoer \cite[p.~48]{Vovk/etal:2015Chervonenkis};
namely, we derive the kernelized versions of the main algorithms of \cite{\OCMXVII}.
In the experimental section (Section~\ref{sec:experiments}),
we demonstrate an important advantage of kernelized versions on an artificial data set.

The standard methods of probabilistic prediction that have been used so far in machine learning,
such as those proposed by Platt \cite{Platt:2000} and Zadrozny and Elkan \cite{Zadrozny/Elkan:2001},
are outside the scope of this paper for two reasons:
first, they have no validity guarantees whatsoever,
and second, they are applicable to classification problems,
whereas in this paper we are interested in regression.
A sister method to conformal prediction, Venn prediction, does have validity guarantees
akin to those in conformal prediction (see, e.g., \cite[Theorems~1 and~2]{\OCMVII}),
but it is also applicable only to classification problems.
Conformalized kernel ridge regression, albeit in the form of prediction intervals rather than predictive distributions,
has been studied by Burnaev and Nazarov~\cite{Burnaev/Nazarov:2016}.

\section{The problem}
\label{sec:setting}

In this section we will introduce our basic prediction problem.
The training sequence consists of $n$ observations $z_i=(x_i,y_i)\in\mathbf{X}\times\mathbf{Y}=\mathbf{X}\times\R$, $i=1,\ldots,n$;
given a test object $x_{n+1}$ we are asked to predict its label $y_{n+1}$.
Each observation $z_i=(x_i,y_i)$, $i=1,\ldots,n+1$, consists of two components,
the object $x_i$ assumed to belong to a measurable space $\mathbf{X}$ that we call the \emph{object space}
and the label $y_i$ that belongs to a measurable space $\mathbf{Y}$ that we call the \emph{label space}.
In this paper we are interested in the case of regression, where the object space is the real line, $\mathbf{Y}=\R$.

In the problem of probability forecasting our prediction takes the form of a probability measure on the label space $\mathbf{Y}$;
since $\mathbf{Y}=\R$, this measure can be represented by its distribution function.
This paper is be devoted to this problem and its modifications.

Our prediction problem can be tackled under different assumptions.
In the chronological order, the standard assumptions are Bayesian (discussed in Section~\ref{sec:Bayesian} below),
statistical parametric (discussed in Section~\ref{sec:fiducial}),
and nonparametric, especially the IID model, standard in machine learning
(and discussed in detail in the rest of this section and further sections).
When using the method of conformal prediction,
it becomes convenient to differentiate between two kinds of assumptions,
hard and soft (to use the terminology of \cite{\OCMI}).
Our hard assumption is the IID model:
the observations are generated independently from the same probability distribution.
The validity of our probabilistic forecasts will depend only on the hard model.
In designing prediction algorithms,
we may also use, formally or informally,
another model in hope that it will be not too far from being correct
and under which we optimize efficiency.
Whereas the hard model is a standard statistical model (the IID model in this paper),
the soft model is not always even formalized;
a typical soft model (avoided in this paper) is the assumption
that the label $y$ of an object $x$ depends on $x$ in an approximately linear fashion.

In the rest of this paper we will use a fixed parameter $a>0$,
determining the amount of regularization that we wish to apply to our solution
to the problem of prediction.
Regularization becomes indispensable when kernel methods are used.

\section{Bayesian solution}
\label{sec:Bayesian}

A very satisfactory solution to our prediction problem
(and plethora of other problems of prediction and inference)
is given by the theory that dominated statistical inference for more than 150 years,
from the work of Thomas Bayes % (nonconformist priest)
and Pierre-Simon Laplace to that of Karl Pearson,
roughly from 1770 to 1930.
This theory, however, requires rather strong assumptions.

Let us assume that our statistical model is linear in a feature space
(спрямляемое пространство, in the terminology of Braverman and his colleagues)
and the noise is Gaussian.
Namely, we assume that $x_1,\ldots,x_{n+1}$ is a deterministic sequence of objects
and that the labels are generated as
\begin{equation}\label{eq:model}
  y_i = w\cdot F(x_i) + \xi_i,
  \quad
  i=1,\ldots,n+1,
\end{equation}
where $F:\mathbf{X}\to H$ is a mapping from the object space to a Hilbert space $H$,
``$\cdot$'' is the dot product in $H$,
$w$ is a random vector distributed as $N(0,(\sigma^2/a)I)$ ($I$ being the identity operator on $H$),
and $\xi_i$ are random variables distributed as $N(0,\sigma^2)$
and independent of $w$ and between themselves.
Here $a$ is the regularization constant introduced at the end of Section~\ref{sec:setting},
and $\sigma>0$ is another parameter, the standard deviation of the noise variables $\xi_i$.

It is easy to check that
\begin{equation}\label{eq:covariance}
  \begin{aligned}
    \Expect y_i &= 0,
    & i=1,\ldots,n,\\
    \cov(y_i,y_j)
    &=
    \frac{\sigma^2}{a}
    \mathcal{K}(x_i,x_j)
    +
    \sigma^2 1_{\{i=j\}},
    & i,j=1,\ldots,n,
  \end{aligned}
\end{equation}
where $\mathcal{K}(x,x'):=F(x)\cdot F(x')$.
By the theorem on normal correlation (see, e.g., \cite[Theorem~II.13.2]{Shiryaev:2004}),
the Bayesian predictive distribution for $y_{n+1}$ given $x_{n+1}$ and the training sequence is
\begin{equation}\label{eq:predictive-dual}
  N
  \left(
    k'(K+aI)^{-1}Y,
    \frac{\sigma^2}{a}\kappa + \sigma^2
    - \frac{\sigma^2}{a} k'(K+aI)^{-1}k
  \right),
\end{equation}
where $k$ is the $n$-vector $k_i:=\mathcal{K}(x_i,x_{n+1})$, $i=1,\ldots,n$,
$K$ is the kernel matrix for the first $n$ observations (the training observations only),
$K_{i,j}:=\mathcal{K}(x_i,x_j)$, $i,j=1,\ldots,n$,
$I=I_n$ is the $n\times n$ unit matrix,
$Y:=(y_1,\ldots,y_n)'$ is the vector of the $n$ training labels,
and $\kappa:=\mathcal{K}(x_{n+1},x_{n+1})$.

The weakness of the model~\eqref{eq:model}
(used, e.g., in \cite[Section~10.3]{Vovk/etal:2005book})
is that the Gaussian measure $N(0,(\sigma^2/a)I)$ exists only when $H$ is finite-dimensional,
but we can circumvent this difficulty
by using \eqref{eq:covariance} directly as our Bayesian model,
for a given symmetric positive semidefinite $\mathcal{K}$.
The mapping $F$ in not part of the picture any longer.
This is the standard approach in Gaussian process regression in machine learning.

In the Bayesian solution, there is no difference between the hard and soft model;
in particular, \eqref{eq:covariance} is required
for the validity of the predictive distribution~\eqref{eq:predictive-dual}.

\section{Fiducial predictive distributions}
\label{sec:fiducial}

After its sesquicentennial rule,
Bayesian statistics was challenged by Fisher and Neyman,
who had little sympathy with each other's views
apart from their common disdain for Bayesian methods.
Fisher's approach was more ambitious, and his goal was to compute a full probability distribution
for a future value (test label in our context) or for the value of a parameter.
Neyman and his followers were content with computing intervals for future values (prediction intervals)
and values of a parameter (confidence intervals).

Fisher and Neyman relaxed the assumptions of Bayesian statistics by allowing uncertainty,
in Knight's \cite{Knight:1921} terminology.
In Bayesian statistics we have an overall probability measure,
i.e., we are in a situation of risk without any uncertainty.
Fisher and Neyman worked in the framework of parametric statistics,
in which we do not have any stochastic model for the value of the parameter
(a number or an element of a Euclidean space).
In the next section we will discuss the next step,
in which the amount of uncertainty (where we lack a stochastic model) is even greater:
our statistical model will be the nonparametric IID model (standard in machine learning).

The available properties of validity naturally become weaker as we weaken our assumptions.
For predicting future values,
conformal prediction (to be discussed in the next section)
ensures calibration in probability, in the terminology of \cite[Definition~1]{Gneiting/Katzfuss:2014}.
It can be shown that Bayesian prediction satisfies a stronger conditional version of this property:
Bayesian predictive distributions are calibrated in probability conditionally on the training sequence and test object
(more generally, on the past).
The property of being calibrated in probability for conformal prediction is, on the other hand,
unconditional; or, in other words, it is conditional on the trivial $\sigma$-algebra.
Fisher's fiducial predictive distributions satisfy an intermediate property of validity:
they are calibrated in probability conditionally
on what was called the $\sigma$-algebra of invariant events in \cite{McCullagh/etal:2009},
which is greater than the trivial $\sigma$-algebra but smaller than the $\sigma$-algebra representing the full knowledge of the past.
Our plan is to give precise statements with proofs in future work.

Fisher did not formalize his fiducial inference, and it has often been regarded as erroneous
(his ``biggest blunder'' \cite{Efron:1998}).
Neyman's simplification,
replacing probability distributions by intervals,
allowed him to state suitable notions of validity more easily,
and his approach to statistics became mainstream
until the Bayesian approach started to reassert itself towards the end of the 20th century.
However, there has been a recent revival of interest in fiducial inference:
cf.\ the BFF (Bayesian, frequentist, and fiducial) series of workshops,
with the fourth one held on 1--3 May 2017 in Cambridge, MA, right after the Braverman Readings in Boston.
Fiducial inference is a key topic of the series,
both in the form of confidence distributions
(the term introduced by David Cox \cite{Cox:1958} in 1958 for distributions for parameters)
and predictive distributions
(which by definition \cite[Definition 1]{Shen/etal:2017} must be calibrated in probability).

Since fiducial inference was developed in the context of parametric statistics,
it has two versions, one targeting computing confidence distributions
and the other predictive distributions.
Under nonparametric assumptions, such as our IID model,
we are not interested in confidence distributions
(the parameter space, the set of all probability measures on the observation space $\mathbf{X}\times\R$,
is just too big),
and concentrate on predictive distributions.
The standard notion of validity for predictive distributions,
introduced independently by Schweder and Hjort \cite[Chapter~12]{Schweder/Hjort:2016}
and Shen, Liu, and Xie \cite{Shen/etal:2017},
is calibration in probability, going back to Philip Dawid's work
(see, e.g., \cite[Section~5.3]{Dawid:1984} and \cite{Dawid/Vovk:1999}).

\section{Conformal predictive distributions}
\label{sec:conformal}

In order to obtain valid predictive distributions under the IID model,
we will need to relax slightly the notion of a predictive distribution
as given in \cite{Shen/etal:2017}.
In our definition we will follow \cite{\OCMXVII} and \cite{\OCMXVIII};
see those papers for further intuition and motivation.

Let $U=U[0,1]$ be the uniform probability distribution on the interval $[0,1]$.
We fix the length $n$ of the training sequence.
Set $\mathbf{Z}:=\mathbf{X}\times\R$; this is our \emph{observation space}.

A function $Q:\mathbf{Z}^{n+1}\times[0,1]\to[0,1]$ is a \emph{randomized predictive system} (RPS) if:
\begin{itemize}
\item[R1a]
  For each training sequence $(z_1,\ldots,z_n)\in\mathbf{Z}^n$ and each test object $x_{n+1}\in\mathbf{X}$,
  the function $Q(z_1,\ldots,z_n,(x_{n+1},y),\tau)$ is monotonically increasing in both $y$ and $\tau$.
\item[R1b]
  For each training sequence $(z_1,\ldots,z_n)\in\mathbf{Z}^n$ and each test object $x_{n+1}\in\mathbf{X}$,
  \begin{align*}
    \lim_{y\to-\infty} Q(z_1,\ldots,z_n,(x_{n+1},y),0) &= 0,\\
    \lim_{y\to\infty} Q(z_1,\ldots,z_n,(x_{n+1},y),1) &= 1.
  \end{align*}
\item[R2]
  For any probability measure $P$ on $\mathbf{Z}$,
  $Q(z_1,\ldots,z_n,z_{n+1},\tau)\sim U$ when $(z_1,\ldots,z_{n+1},\tau)\sim P^{n+1}\times U$.
\end{itemize}
The function
\begin{equation}\label{eq:Q-n}
  Q_n:(y,\tau)\in\R\times[0,1]\mapsto Q(z_1,\ldots,z_n,(x_{n+1},y),\tau)
\end{equation}
is the \emph{randomized predictive distribution (function)} (RPD)
output by the randomized predictive system $Q$ on a training sequence $z_1,\ldots,z_n$ and a test object $x_{n+1}$.

A \emph{conformity measure} is a measurable function $A:\mathbf{Z}^{n+1}\to\R$
that is invariant with respect to permutations of the first $n$ observations.
A simple example, used in this paper, is
\begin{equation}\label{eq:example}
  A(z_1,\ldots,z_{n+1})
  :=
  y_{n+1}-\hat y_{n+1},
\end{equation}
$\hat y_{n+1}$ being the prediction for $y_{n+1}$ computed from $x_{n+1}$
and $z_1,\ldots,z_{n+1}$ as training sequence.
The \emph{conformal transducer} determined by a conformity measure $A$ is defined as
\begin{multline}\label{eq:conformal-Q}
  Q(z_1,\ldots,z_n,(x_{n+1},y),\tau)
  :=
  \frac{1}{n+1}
  \Bigl(
    \left|\left\{i=1,\ldots,n+1\mid\alpha^y_i<\alpha^y_{n+1}\right\}\right|\\
    +
    \tau
    \left|\left\{i=1,\ldots,n+1\mid\alpha^y_i=\alpha^y_{n+1}\right\}\right|
  \Bigr),
\end{multline}
where $(z_1,\ldots,z_n)\in\mathbf{Z}^n$ is a training sequence,
$x_{n+1}\in\mathbf{X}$ is a test object,
and for each $y\in\R$ the corresponding \emph{conformity scores} $\alpha_i^y$ are defined by
\begin{equation}\label{eq:conformity-scores}
  \begin{aligned}
    \alpha_i^y
    &:=
    A(z_1,\ldots,z_{i-1},z_{i+1},\ldots,z_n,(x_{n+1},y),z_i),
      \qquad i=1,\ldots,n,\\
    \alpha^y_{n+1}
    &:=
    A(z_1,\ldots,z_n,(x_{n+1},y)).
  \end{aligned}
\end{equation}
A function is a \emph{conformal transducer} if it is the conformal transducer determined by some conformity measure.
A \emph{conformal predictive system} (CPS) is a function which is both a conformal transducer
and a randomized predictive system.
A \emph{conformal predictive distribution} (CPD) is a function $Q_n$ defined by \eqref{eq:Q-n}
for a conformal predictive system $Q$.

The following lemma, stated in \cite{\OCMXVII}, gives simple conditions for a conformal transducer to be an RPS;
it uses the notation of~\eqref{eq:conformity-scores}.

\begin{lemma}\label{lem:criterion}
  The conformal transducer determined by a conformity measure $A$ is an RPS if,
  for each training sequence $(z_1,\ldots,z_n)\in\mathbf{Z}^n$,
  each test object $x_{n+1}\in\mathbf{X}$,
  and each $i\in\{1,\ldots,n\}$:
  \begin{itemize}
  \item
    $\alpha^y_{n+1}-\alpha_i^y$ is a monotonically increasing function of $y\in\R$;
  \item
    $
      \lim_{y\to\pm\infty}
      \left(
        \alpha^y_{n+1}-\alpha_i^y
      \right)
      =
      \pm\infty
    $.
  \end{itemize}
\end{lemma}

\section{Kernel Ridge Regression Prediction Machine}

In this section we introduce the Kernel Ridge Regression Prediction Machine (KRRPM);
it will be the conformal transducer determined by a conformity measure of the form \eqref{eq:example},
where $\hat y_{n+1}$ is computed using kernel ridge regression, to be defined momentarily.
There are three natural versions of the definition, and we start from reviewing them.
All three versions are based on \eqref{eq:model} as soft model (with the IID model being the hard model).

Given a training sequence $(z_1,\ldots,z_n)\in\mathbf{Z}^n$ and a test object $x_{n+1}\in\mathbf{X}$,
the \emph{kernel ridge regression} predicts
\[
  \hat y_{n+1}
  :=
  k'(K+aI)^{-1}Y
\]
for the label $y_{n+1}$ of $x_{n+1}$.
This is just the mean in \eqref{eq:predictive-dual}, and the variance is ignored.
Plugging this definition into~\eqref{eq:example},
we obtain the \emph{deleted KRRPM}.
Alternatively, we can replace the conformity measure~\eqref{eq:example}
by
\begin{equation}\label{eq:example-bis}
  A(z_1,\ldots,z_{n+1})
  :=
  y_{n+1}-\hatbar y_{n+1},
\end{equation}
where
\begin{equation}\label{eq:notation}
  \hatbar y_{n+1}
  :=
  \bar k'(\bar K+aI)^{-1}\bar Y
\end{equation}
is the prediction for the label $y_{n+1}$ of $x_{n+1}$
computed using $z_1,\ldots,z_{n+1}$ as the training sequence.
The notation used in~\eqref{eq:notation} is:
$\bar k$ is the $(n+1)$-vector $k_i:=\mathcal{K}(x_i,x_{n+1})$, $i=1,\ldots,n+1$,
$\bar K$ is the kernel matrix for all $n+1$ observations,
$\bar K_{i,j}:=\mathcal{K}(x_i,x_j)$, $i,j=1,\ldots,n+1$,
$I=I_{n+1}$ is the $(n+1)\times(n+1)$ unit matrix,
and $\bar Y:=(y_1,\ldots,y_{n+1})'$ is the vector of all $n+1$ labels.
In this context, $\mathcal{K}$ is any given \emph{kernel},
i.e., symmetric positive semidefinite function $\mathcal{K}:\mathbf{X}^2\to\R$.
The corresponding conformal transducer is the \emph{ordinary KRRPM}.
The disadvantage of the deleted and ordinary KRRPM is that they are not RPSs
(they can fail to produce a function increasing in $y$
in the presence of extremely high-leverage objects).

Set
\begin{equation}\label{eq:hat}
  \bar H
  :=
  (\bar K+aI)^{-1}\bar K
  =
  \bar K(\bar K+aI)^{-1}.
\end{equation}
This \emph{hat matrix} ``puts hats on the $y$s'':
according to~\eqref{eq:notation}, $\bar H\bar Y$ is the vector $(\hatbar y_1,\ldots,\hatbar y_{n+1})'$,
where $\hatbar y_i$, $i=1,\ldots,n+1$,
is the prediction for the label $y_i$ of $x_i$ computed using $z_1,\ldots,z_{n+1}$ as the training sequence.
We will refer to the entries of the matrix $\bar H$ as $\bar h_{i,j}$
(where $i$ is the row and $j$ is the column of the entry),
abbreviating $\bar h_{i,i}$ to $\bar h_i$.
The usual relation between the residuals in~\eqref{eq:example} and~\eqref{eq:example-bis} is
\begin{equation}\label{eq:relation}
  y_{n+1}-\hat y_{n+1}
  =
  \frac{y_{n+1}-\hatbar y_{n+1}}{1-\bar h_{n+1}}.
\end{equation}
This equality makes sense since the diagonal elements $\bar h_i$ of the hat matrix are always in the semi-open interval $[0,1)$
(and so the numerator is non-zero); for details, see Appendix~A.
Equation~\eqref{eq:relation} motivates using the \emph{studentized residuals} $(y_{n+1}-\hatbar y_{n+1})(1-\bar h_{n+1})^{-1/2}$,
which are half-way between the deleted residuals in~\eqref{eq:example}
and the ordinary residuals in~\eqref{eq:example-bis}.
(We ignore a factor in the usual definition of studentized residuals, as in \cite[(4.8)]{Montgomery/etal:2012},
that does not affect the value \eqref{eq:conformal-Q} of the conformal transducer.)
The conformal transducer determined by the corresponding conformity measure
\begin{equation}\label{eq:Studentized-conformity}
  A(z_1,\ldots,z_{n+1})
  :=
  \frac{y_{n+1}-\hatbar y_{n+1}}{\sqrt{1-\bar h_{n+1}}}
\end{equation}
is the (studentized) \emph{KRRPM}.
Later in this section we will see that the KRRPM is an RPS.
This is the main reason why this is the main version considered in this paper, with ``studentized'' usually omitted.

\subsection*{An explicit form of the KRRPM}

According to \eqref{eq:conformal-Q},
to compute the predictive distributions produced by the KRRPM
(in its studentized version),
we need to solve the equation $\alpha_i^y=\alpha_{n+1}^y$
(and the corresponding inequality $\alpha_i^y<\alpha_{n+1}^y$)
for $i=1,\ldots,n+1$.
Combining the definition~\eqref{eq:conformity-scores} of the conformity scores $\alpha_i^y$,
the definition~\eqref{eq:Studentized-conformity} of the conformity measure,
and the fact that the predictions $\hatbar y_i$ can be obtained from $\bar Y$ by applying the hat matrix $\bar H$ (cf.~\eqref{eq:hat}),
we can rewrite $\alpha_i^y=\alpha_{n+1}^y$ as
\[
  \frac{y_i-\sum_{j=1}^n\bar h_{ij}y_j-\bar h_{i,n+1}y}{\sqrt{1-\bar h_i}}
  =
  \frac{y-\sum_{j=1}^n\bar h_{n+1,j}y_j-\bar h_{n+1}y}{\sqrt{1-\bar h_{n+1}}}.
\]
This is a linear equation, $A_i=B_i y$, and solving it we obtain $y=C_i:=A_i/B_i$, where
\begin{align}
  A_i
  &:=
  \frac{\sum_{j=1}^n\bar h_{n+1,j}y_j}{\sqrt{1-\bar h_{n+1}}}
  +
  \frac{y_i-\sum_{j=1}^n\bar h_{ij}y_j}{\sqrt{1-\bar h_i}},
  \label{eq:A}\\
  B_i
  &:=
  \sqrt{1-\bar h_{n+1}}
  +
  \frac{\bar h_{i,n+1}}{\sqrt{1-\bar h_i}}.
  \label{eq:B}
\end{align}
The following lemma, to be proved in Appendix~A, allows us to compute~\eqref{eq:conformal-Q} easily.
\begin{lemma}\label{lem:B}
  It is always true that $B_i>0$.
\end{lemma}
The lemma gives Algorithm~\ref{alg:KRRPM} for computing the conformal predictive distribution~\eqref{eq:Q-n}.
The notation $i'$ and $i''$ used in line~\ref{ln:notation} is defined as
$i':=\min\{j\mid C_{(j)}=C_{(i)}\}$ and $i'':=\max\{j\mid C_{(j)}=C_{(i)}\}$,
to ensure that $Q_n(y,0)=Q_n(y-,0)$ and $Q_n(y,1)=Q_n(y+,1)$ at $y=C_{(i)}$;
$C_{(0)}$ and $C_{(n+1)}$ are understood to be $-\infty$ and $\infty$, respectively.
Notice that there is no need to apply Lemma~\ref{lem:criterion} formally;
Lemma~\ref{lem:B} makes it obvious that the KRRPM is a CPS.

\begin{algorithm}[bt]
  \caption{Kernel Ridge Regression Prediction Machine}
  \label{alg:KRRPM}
  \begin{algorithmic}[1]
    \Require
      A training sequence $(x_i,y_i)\in\mathbf{X}\times\R$, $i=1,\ldots,n$.
    \Require
      A test object $x_{n+1}\in\mathbf{X}$.
    \State Define the hat matrix $\bar H$ by \eqref{eq:hat}, $\bar K$ being the $(n+1)\times(n+1)$ kernel matrix.
    \For{$i\in\{1,2,\ldots,n\}$}
      \State Define $A_i$ and $B_i$ by \eqref{eq:A} and \eqref{eq:B}, respectively.
      \State Set $C_i:=A_i/B_i$.
    \EndFor
    \State Sort $C_1,\ldots,C_n$ in the increasing order obtaining $C_{(1)}\le\cdots\le C_{(n)}$.
    \State Return the following predictive distribution for $y_{n+1}$:\label{ln:notation}
      \begin{equation}\label{eq:return}
        Q_n(y,\tau)
        :=
        \begin{cases}
          \frac{i+\tau}{n+1} & \text{if $y\in(C_{(i)},C_{(i+1)})$ for $i\in\{0,1,\ldots,n\}$}\\
          \frac{i'-1+\tau(i''-i'+2)}{n+1} & \text{if $y=C_{(i)}$ for $i\in\{1,\ldots,n\}$}.
        \end{cases}
      \end{equation}
  \end{algorithmic}
\end{algorithm}

Algorithm~\ref{alg:KRRPM} is not computationally efficient for a large test set,
since the hat matrix $\bar H$ (cf.\ \eqref{eq:hat}) needs to be computed from scratch for each test object.
To obtain a more efficient version,
we use a standard formula for inverting partitioned matrices
(see, e.g., \cite[(8)]{Henderson/Searle:1981} or \cite[(2.44)]{Vovk/etal:2005book})
to obtain
\begin{align}
  \bar H
  &=
  (\bar K+aI)^{-1}\bar K
  =
  \begin{pmatrix}
    K+aI & k \notag\\
    k' & \kappa+a
  \end{pmatrix}^{-1}
  \begin{pmatrix}
    K & k \\
    k' & \kappa
  \end{pmatrix}\notag\\
  &=
  \begin{pmatrix}
    (K+aI)^{-1} + d (K+aI)^{-1} k k' (K+aI)^{-1}
    & -d (K+aI)^{-1} k \\
    -d k' (K+aI)^{-1}
    & d
  \end{pmatrix}
  \begin{pmatrix}
    K & k \\
    k' & \kappa
  \end{pmatrix}\notag\\
  &=
  \left(
    \begin{matrix}
      H+d(K+aI)^{-1}k k' H - d(K+aI)^{-1} k k'\\
      -d k' H + d k'
    \end{matrix}
  \right.\label{eq:matrix-1}\\
  &\qquad
  \left.
    \begin{matrix}
      (K+aI)^{-1}k + d(K+aI)^{-1} k k' (K+aI)^{-1}k - d\kappa(K+aI)^{-1}k\\
      -d k'(K+aI)^{-1}k + d\kappa
    \end{matrix}
  \right)\label{eq:matrix-2}\\
  &=
  \begin{pmatrix}
    H+d(K+aI)^{-1}k k' (H-I) & d(I-H)k\\
    d k'(I-H) & -d k'(K+aI)^{-1}k + d\kappa
  \end{pmatrix}\label{eq:final-matrix}\\
  &=
  \begin{pmatrix}
    H-ad(K+aI)^{-1}k k'(K+aI)^{-1} & ad(K+aI)^{-1}k\\
    ad k'(K+aI)^{-1} & d\kappa - d k'(K+aI)^{-1}k
  \end{pmatrix}\label{eq:really-final},
\end{align}
where
\begin{equation}\label{eq:d}
  d
  :=
  \frac{1}{\kappa+a-k'(K+aI)^{-1}k}
\end{equation}
(the denominator is positive by the theorem on normal correlation, already used in Section~\ref{sec:Bayesian}),
the equality in line~\eqref{eq:final-matrix} follows from $\bar H$ being symmetric
(which allows us to ignore the upper right block of the matrix \eqref{eq:matrix-1}--\eqref{eq:matrix-2}),
and the equality in line~\eqref{eq:really-final} follows from
\[
  I-H
  =
 (K+aI)^{-1} (K+aI)
 -
 (K+aI)^{-1} K
  =
  a
  (K+aI)^{-1}.
\]
We have been using the notation $H$ for the training hat matrix
\begin{equation}\label{eq:H}
  H=(K+aI)^{-1}K=K(K+aI)^{-1}.
\end{equation}
Notice that the constant $ad$ occurring in several places in~\eqref{eq:really-final}
is between 0 and~1:
\begin{equation}\label{eq:ad}
  ad
  =
  \frac{a}{a+\kappa-k'(K+aI)^{-1}k}
  \in
  (0,1]
\end{equation}
(the fact that $\kappa-k'(K+aI)^{-1}k$ is nonnegative
follows from the lower right entry $\bar h_{n+1}$ of the hat matrix \eqref{eq:really-final} being nonnegative;
the nonnegativity of the diagonal entries of hat matrices is discussed in Appendix~\ref{app:hat}).

The important components in the expressions for $A_i$ and $B_i$ (cf.\ \eqref{eq:A} and \eqref{eq:B}) are,
according to \eqref{eq:really-final},
\begin{align}
  1-\bar h_{n+1}
  &=
  1 + d k'(K+aI)^{-1}k - d\kappa
  =
  1 + \frac{k'(K+aI)^{-1}k - \kappa}{\kappa+a-k'(K+aI)^{-1}k}\notag\\
  &=
  \frac{a}{\kappa+a-k'(K+aI)^{-1}k}
  =
  ad,
  \label{eq:1-h}\\
  1-\bar h_i
  &=
  1-h_i
  +
  ad
  e'_i (K+aI)^{-1} k
  k' (K+aI) e_i\notag\\
  &=
  1-h_i
  +
  ad
  (e'_i (K+aI)^{-1} k)^2,
  \label{eq:1-h-i}
\end{align}
where $h_i=h_{i,i}$ is the $i$th diagonal entry of the hat matrix \eqref{eq:H}
for the $n$ training objects
and $e_i$ is the $i$th vector in the standard basis of $\mathbb{R}^n$
(so that the $j$th component of $e_i$ is $1_{\{i=j\}}$ for $j=1,\ldots,n$).
Let $\hat y_i:=e'_i H Y$ be the prediction for $y_i$
computed from the training sequence $z_1,\ldots,z_n$ and the test object $x_i$.
Using \eqref{eq:1-h} (but not using \eqref{eq:1-h-i} for now),
we can transform \eqref{eq:A} and \eqref{eq:B} as
\begin{align}
  A_i
  &:=
  \frac{\sum_{j=1}^n\bar h_{n+1,j}y_j}{\sqrt{1-\bar h_{n+1}}}
  +
  \frac{y_i-\sum_{j=1}^n\bar h_{ij}y_j}{\sqrt{1-\bar h_i}}\notag\\
  &=
  (ad)^{-1/2}
  \sum_{j=1}^n ad y_j k'(K+aI)^{-1} e_j\notag\\
  &\quad{}+
  \frac{y_i-\sum_{j=1}^n h_{ij}y_j+\sum_{j=1}^n ad y_j e'_i(K+aI)^{-1} k k' (K+aI)^{-1} e_j}{\sqrt{1-\bar h_i}}\notag\\
  &=
  (ad)^{1/2}
  k'(K+aI)^{-1} Y
  +
  \frac{y_i - \hat y_i + ad e'_i(K+aI)^{-1} k k' (K+aI)^{-1} Y}{\sqrt{1-\bar h_i}},\notag\\
  &=
  \sqrt{ad}
  \hat y_{n+1}
  +
  \frac{y_i - \hat y_i + ad \hat y_{n+1} e'_i(K+aI)^{-1} k}{\sqrt{1-\bar h_i}},
  \label{eq:A-final}
\end{align}
where $\hat y_{n+1}$ is the Bayesian prediction for $y_{n+1}$
(cf.\ the expected value in~\eqref{eq:predictive-dual}),
and
\begin{equation}\label{eq:B-final}
  B_i
  :=
  \sqrt{1-\bar h_{n+1}}
  +
  \frac{\bar h_{i,n+1}}{\sqrt{1-\bar h_i}}
  =
  \sqrt{ad}
  +
  \frac{ad k' (K+aI)^{-1} e_i}{\sqrt{1-\bar h_i}}.
\end{equation}

Therefore, we can implement Algorithm~\ref{alg:KRRPM} as follows.
Preprocessing the training sequence takes time $O(n^3)$
(or faster if using, say, the Coppersmith--Winograd algorithm and its versions;
we assume that the kernel $\mathcal{K}$ can be computed in time $O(1)$):
\begin{enumerate}
\item
  The $n\times n$ kernel matrix $K$ can be computed in time $O(n^2)$.
\item
  The matrix $(K+aI)^{-1}$ can be computed in time $O(n^3)$.
\item
  The diagonal of the training hat matrix $H:=(K+aI)^{-1}K$ can be computed in time $O(n^2)$.
\item
  All $\hat y_i$, $i=1,\ldots,n$, can be computed by $\hat y:=HY=(K+aI)^{-1}(K Y)$ in time $O(n^2)$
  (even without knowing $H$).
\end{enumerate}
Processing each test object $x_{n+1}$ takes time $O(n^2)$:
\begin{enumerate}
\item
  Vector $k$ and number $\kappa$ (as defined after \eqref{eq:predictive-dual})
  can be computed in time $O(n)$ and $O(1)$, respectively.
\item\label{it:Kk}
  Vector $(K+aI)^{-1}k$ can be computed in time $O(n^2)$.
\item
  Number $k'(K+aI)^{-1}k$ can now be computed in time $O(n)$.
\item
  Number $d$ defined by \eqref{eq:d} can be computed in time $O(1)$.
\item
  For all $i=1,\ldots,n$,
  compute $1-\bar h_i$ as \eqref{eq:1-h-i},
  in time $O(n)$ overall
  (given the vector computed in~\ref{it:Kk}).
\item
  Compute the number $\hat y_{n+1}:=k'(K+aI)^{-1}Y$ in time $O(n)$
  (given the vector computed in~\ref{it:Kk}).
\item
  Finally, compute $A_i$ and $B_i$ for all $i=1,\ldots,n$
  as per \eqref{eq:A-final} and \eqref{eq:B-final},
  set $C_i:=A_i/B_i$,
  and output the predictive distribution~\eqref{eq:return}.
  This takes time $O(n)$ except for sorting the $C_i$,
  which takes time $O(n\log n)$.
\end{enumerate}

\section{Limitation of the KRRPM}

The KRRPM makes a significant step forward as compared to the LSPM of \cite{\OCMXVII}:
our soft model \eqref{eq:model} is no longer linear in $x_i$.
In fact, using a universal kernel (such as Laplacian in Section~\ref{sec:experiments})
allows the function $x\in\mathbf{X}\mapsto w\cdot F(x)$ to approximate any continuous function
(arbitrarily well within any compact set in $\mathbf{X}$).
However, since we are interested in predictive distributions rather than point predictions,
using the soft model \eqref{eq:model} still results in the KRRPM being restricted.
In this section we discuss the nature of the restriction,
using the ordinary KRRPM as a technical tool.

The Bayesian predictive distribution \eqref{eq:predictive-dual} is Gaussian and
(as clear from \eqref{eq:model} and from the bottom right entry of \eqref{eq:really-final} being nonnegative)
its variance is at least $\sigma^2$.
We will see that the situation with the conformal distribution is not as bad, despite the remaining restriction.
To understand the nature of the restriction it will be convenient to ignore the denominator in \eqref{eq:Studentized-conformity},
i.e., to consider the ordinary KRRPM;
the difference between the (studentized) KRRPM and ordinary KRRPM will be small
in the absence of high-leverage objects
(an example will be given in the next section).
For the ordinary KRRPM we have, in place of \eqref{eq:A} and \eqref{eq:B},
\begin{align*}
  A_i
  &:=
  \sum_{j=1}^n \bar h_{n+1,j}y_j
  +
  y_i-\sum_{j=1}^n \bar h_{i,j}y_j,\\
  B_i
  &:=
  1-\bar h_{n+1}
  +
  \bar h_{i,n+1}.
\end{align*}
Therefore, \eqref{eq:A-final} and~\eqref{eq:B-final} become
\begin{equation*}
  A_i
  =
  ad
  \hat y_{n+1}
  +
  y_i - \hat y_i
  +
  ad
  \hat y_{n+1}
  e'_i(K+aI)^{-1} k
\end{equation*}
and
\begin{equation*}
  B_i
  =
  ad + ad e'_i(K+aI)^{-1}k,
\end{equation*}
respectively.
For $C_i:=A_i/B_i$ we now obtain
\begin{multline}\label{eq:to-discuss}
  C_i
  =
  \hat y_{n+1}
  +
  \frac{y_i - \hat y_i}{ad + ad e'_i(K+aI)^{-1}k}\\
  =
  \hat y_{n+1}
  +
  \frac{\sigma^2_{{\rm Bayes}}/\sigma^2}{1 + e'_i(K+aI)^{-1}k}
  (y_i - \hat y_i),
\end{multline}
where $\hat y_{n+1}$ is, as before, the Bayesian prediction for $y_{n+1}$,
and $\sigma^2_{{\rm Bayes}}$ is the variance of the Bayesian predictive distribution~\eqref{eq:predictive-dual}
(cf.\ \eqref{eq:ad}).

The second addend $e'_i(K+aI)^{-1}k$ in the denominator of~\eqref{eq:to-discuss}
is the prediction for the label of the test object $x_{n+1}$ in the situation where all training labels are $0$ apart from the $i$th,
which is $1$.
For a long training sequence we can expect it to be close to 0 (unless $x_i$ or $x_{n+1}$ are highly influential);
therefore, we can expect the shape of the predictive distribution output by the ordinary KRRPM
to be similar to the shape of the empirical distribution function of the residuals $y_i-\hat y_i$.
In particular, this shape does not depend (or depends weakly) on the test object $x_{n+1}$.
This lack of sensitivity of the predictive distribution to the test object prevents
the conformal predictive distributions output by the KRRPM from being universally consistent
in the sense of \cite{\OCMXVIII}.
The shape of the predictive distribution can be arbitrary, not necessarily Gaussian (as in~\eqref{eq:predictive-dual}),
but it is fitted to all training residuals and not just the residuals
for objects similar to the test object.
One possible way to get universally consistent conformal predictive distributions
would be to replace the right-hand side of \eqref{eq:example} by $\hat F_{n+1}(y_{n+1})$,
where $\hat F_{n+1}$ is the Bayesian predictive distribution for $y_{n+1}$
computed from $x_{n+1}$ and $z_1,\ldots,z_{n+1}$ as training sequence
for a sufficiently flexible Bayesian model
(in any case, more flexible than our homoscedastic model \eqref{eq:model}).
This idea was referred to as de-Bayesing in \cite[Section~4.2]{Vovk/etal:2005book}
and frequentizing in \cite[Section~3]{Wasserman:2011}.
However, modelling input-dependent (heteroscedastic) noise efficiently is a well-known difficult problem
in Bayesian regression, including Gaussian process regression
(see, e.g., \cite{Goldberg/etal:1998,Le/etal:2005,Snelson/Ghahramani:2006}).

\section{Experimental results}
\label{sec:experiments}

In the first part of this section we illustrate the main advantage of the KRRPM over the LSPM introduced in \cite{\OCMXVII},
its flexibility:
for a suitable kernel,
it gets the location of the predictive distribution right.
In the second part, we illustrate the limitation discussed in the previous section:
while the KRRPM adapts to the shape of the distribution of labels,
the adaptation is not conditional on the test object.
Both points will be demonstrated using artificial data sets.

In our first experiment we generate a training sequence of length 1000 from the model
\begin{equation}\label{eq:truth}
  y_i
  =
  w_1 \cos x_{i,1} + w_2 \cos x_{i,2}
  +
  w_3 \sin x_{i,1} + w_4 \sin x_{i,2}
  +
  \xi_i,
\end{equation}
where $(w_1,w_2,w_3,w_4)\sim N(0,I_4)$ ($I_4$ being the unit $4\times4$ matrix),
$(x_{i,1},x_{i,2})\sim U[-1,1]^2$ ($U[-1,1]$ being the uniform probability distribution on $[-1,1]$),
and $\xi_i\sim N(0,1)$, all independent.
This corresponds to the Bayesian ridge regression model with $a=\sigma=1$.
The true kernel is
\begin{align}
  &\mathcal{K}((x_1,x_2),(x'_1,x'_2))\notag\\
  &=
  (\cos x_1,\cos x_2,\sin x_1,\sin x_2)
  \cdot
  (\cos x'_1,\cos x'_2,\sin x'_1,\sin x'_2)\notag\\
  &=
  \cos(x_1-x'_1) + \cos(x_2-x'_2).
  \label{eq:correct}
\end{align}

Remember that a kernel is \emph{universal} \cite{Steinwart:2001}
if any continuous function can be uniformly approximated (over each compact set)
by functions in the corresponding reproducing kernel Hilbert space.
An example of a universal kernel is the \emph{Laplacian kernel}
\[
  \mathcal{K}(x,x')
  :=
  \exp
  \left(
    -\left\|x-x'\right\|
  \right).
\]
Laplacian kernels were introduced and studied in \cite{Thomas-Agnan:1996};
the corresponding reproducing kernel Hilbert space has the Sobolev norm
\[
  \left\|u\right\|^2
  =
  2
  \int_{-\infty}^{\infty} u(t)^2 \dd t
  +
  2
  \int_{-\infty}^{\infty} u'(t)^2 \dd t
\]
(see \cite[Corollary~1]{Thomas-Agnan:1996}).
This expression shows that Laplacian kernels are indeed universal.
On the other hand, the \emph{linear kernel} $\mathcal{K}(x,x'):=x\cdot x'$
is far from being universal;
remember that the LSPM \cite{\OCMXVII} corresponds to this kernel and $a=0$.

\begin{figure}[tb]
  \begin{center}
    \includegraphics[width=0.3\columnwidth]{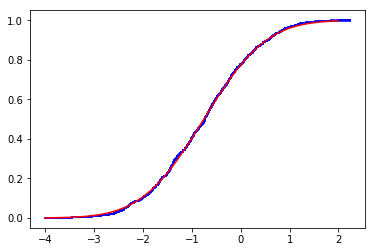}
    \hfil
    \includegraphics[width=0.3\columnwidth]{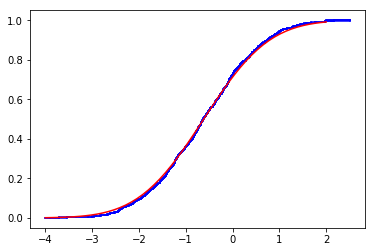}
    \hfil
    \includegraphics[width=0.3\columnwidth]{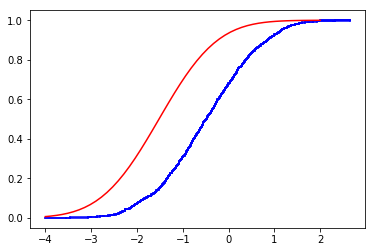}
  \end{center}
  \caption{The predictive distribution for the label of the test object $(1,1)$
    based on a training sequence of length 1000 (all generated from the model \eqref{eq:truth}).
    The red line in each panel is the Bayesian predictive distribution based on the true kernel~\eqref{eq:correct},
    and the blue line is the conformal predictive distribution based on:
    the true kernel~\eqref{eq:correct} in the left-most panel;
    the Laplacian kernel in the middle panel;
    the linear kernel in the right-most panel.}
  \label{fig:Bayes_vs}
\end{figure}

Figure~\ref{fig:Bayes_vs} shows that, on this data set, universal kernels lead to better results.
The parameter $a$ in Figure~\ref{fig:Bayes_vs} is the true one, $a=1$.
In the case of the Bayesian predictive distribution,
the parameter $\sigma=1$ is also the true one;
remember that conformal predictive distributions do not require $\sigma$.
The right-most panel shows that, when based on the linear kernel,
the conformal predictive distribution can get the predictive distribution wrong.
The other two panels show that the true kernel and, more importantly, the Laplacian kernel
(chosen independently of the model \eqref{eq:truth}) are much more accurate.
Figure~\ref{fig:Bayes_vs} shows predictive distributions for a specific test object, $(1,1)$,
but this behaviour is typical.

\begin{figure}[tb]
  \begin{center}
    \includegraphics[width=0.45\columnwidth]{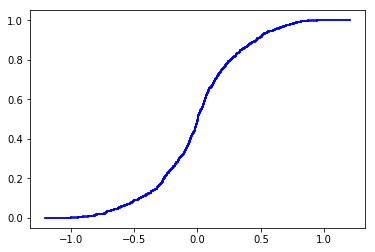}
    \hfil
    \includegraphics[width=0.45\columnwidth]{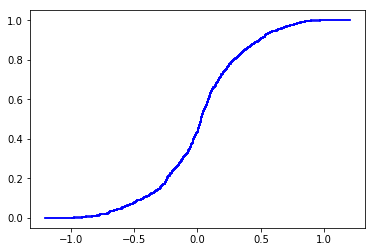}
  \end{center}
  \caption{Left panel: predictions of the KRRPM for a training sequence of length $1000$ and $x_{1001}=0$.
    Right panel: predictions for $x_{1001}=1$.
    The data are described in the text.}
    \label{fig:insensitive_1000}
\end{figure}

We now illustrate the limitation of the KRRPM that we discussed in the previous section.
An artificial data set is generated as follows:
$x_i\in[0,1]$, $i=1,\ldots,n$, are chosen independently from the uniform distribution $U$ on $[0,1]$,
and $y_i\in[-x_i,x_i]$ are then chosen independently, again from the uniform distributions $U[-x_i,x_i]$ on their intervals.
Figure~\ref{fig:insensitive_1000} shows the prediction for $x_{n+1}=0$ on the left and for $x_{n+1}=1$ on the right
for $n=1000$; there is no visible difference between the studentized and ordinary versions of the KRRPM.
The difference between the predictions for $x_{n+1}=0$ and $x_{n+1}=1$ is slight,
whereas ideally we would like the former prediction to be concentrated at 0
whereas the latter should be close to the uniform distribution on $[-1,1]$.

\begin{figure}[tb]
  \begin{center}
    \includegraphics[width=0.45\columnwidth]{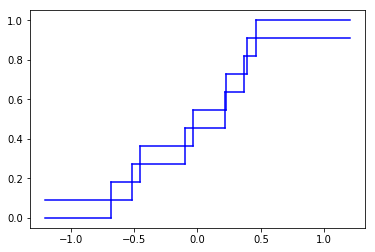}
    \hfil
    \includegraphics[width=0.45\columnwidth]{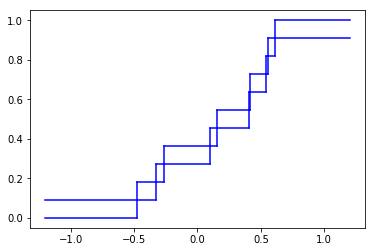}\\
    \includegraphics[width=0.45\columnwidth]{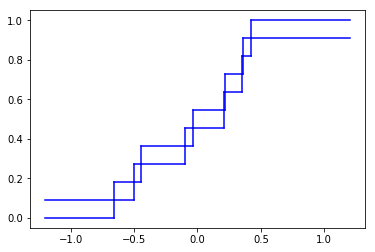}
    \hfil
    \includegraphics[width=0.45\columnwidth]{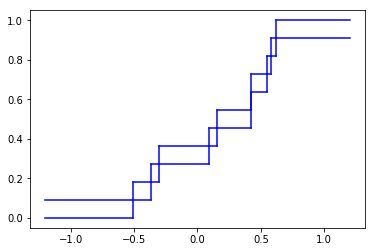}
  \end{center}
  \caption{Upper left panel: predictions of the (studentized) KRRPM for a training sequence of length $10$ and $x_{11}=0$.
    Upper right panel: analogous predictions for $x_{11}=1$.
    Lower left panel: predictions of the ordinary KRRPM for a training sequence of length $10$ and $x_{11}=0$.
    Lower right panel: analogous predictions for $x_{11}=1$.}
    \label{fig:insensitive_10}
\end{figure}

Fine details can be seen in Figure~\ref{fig:insensitive_10},
which is analogous to Figure~\ref{fig:insensitive_1000} but uses a training sequence of length $n=10$.
It shows the plots of the functions $Q_n(y,0)$ and $Q_n(y,1)$ of $y$,
in the notation of \eqref{eq:Q-n}.
These functions carry all information about $Q_n(y,\tau)$ as function of $y$ and $\tau$
since $Q_n(y,\tau)$ can be computed as the convex mixture $(1-\tau)Q_n(y,0)+\tau Q_n(y,1)$ of $Q_n(y,0)$ and $Q_n(y,1)$.

In all experiments described in this section,
the seed of the Python pseudorandom numbers generator was set to 0 for reproducibility.
Our code can be found at \href{http://alrw.net}{http://alrw.net} (Working Paper~20).

\subsection*{Acknowledgments}

This work has been supported by the EU Horizon 2020 Research and Innovation programme
(in the framework of the ExCAPE project under grant agreement 671555)
and Astra Zeneca
(in the framework of the project ``Machine Learning for Chemical Synthesis'').

\appendix
\section{Properties of the hat matrix}
\label{app:hat}

In the kernelized setting of this paper the hat matrix is defined as $H=(K+aI)^{-1}K$,
where $K$ is a symmetric positive semidefinite matrix whose size is denoted $n\times n$ in this appendix
(cf.~\eqref{eq:hat}; in our current abstract setting we drop the bars over $H$ and $K$
and write $n$ in place of $n+1$).
We will prove, or give references for, various properties of the hat matrix used in the main part of the paper.

Numerous useful properties of the hat matrix can be found in literature
(see, e.g., \cite{Chatterjee/Hadi:1988}).
However, the usual definition of the hat matrix is different from ours,
since it is not kernelized;
therefore, we start from reducing our kernelized definition to the standard one.
Since $K$ is symmetric positive semidefinite,
it can be represented in the form $K=XX'$ for some matrix $X$, whose size will be denoted $n\times p$
(in fact, a matrix is symmetric positive semidefinite if and only if it can be represented as the Gram matrix
of $n$ vectors;
this easily follows from the fact that a symmetric positive semidefinite $K$ can be diagonalized: $K=Q'\Lambda Q$,
where $Q$ and $\Lambda$ are $n\times n$ matrices, $\Lambda$ is diagonal with nonnegative entries, and $Q'Q=I$).
Now we can transform the hat matrix as
\[
  H
  =
  (K+aI)^{-1}K
  =
  (XX'+aI)^{-1}XX'
  =
  X(X'X+aI)^{-1}X'
\]
(the last equality can be checked by multiplying both sides by $(XX'+aI)$ on the left).
If we now extend $X$ by adding $\sqrt{a}I_p$ on top of it
(where $I_p=I$ is the $p\times p$ unit matrix),
\begin{equation}\label{eq:extension}
  \tilde X
  :=
  \begin{pmatrix}
    \sqrt{a}I_p \\ X
  \end{pmatrix},
\end{equation}
and set
\begin{equation}\label{eq:H-tilde}
  \tilde H
  :=
  \tilde X(\tilde X'\tilde X)^{-1}\tilde X'
  =
  \tilde X(X'X+aI)^{-1}\tilde X',
\end{equation}
we will obtain a $(p+n)\times(p+n)$ matrix containing $H$ in its lower right $n\times n$ corner.
To find $HY$ for a vector $Y\in\R^n$, we can extend $Y$ to $\tilde Y\in\R^{p+n}$
by adding $p$ zeros at the beginning of $Y$ and then discard the first $p$ elements in $\tilde H \tilde Y$.
Notice that $\tilde H$ is the usual definition of the hat matrix associated with the data matrix $\tilde X$
(cf.\ \cite[(1.4a)]{Chatterjee/Hadi:1988}).

When discussing \eqref{eq:relation}, we used the fact that the diagonal elements of $H$ are in $[0,1)$.
It is well-known that the diagonal elements of the usual hat matrix, such as $\tilde H$,
are in $[0,1]$ (see, e.g., \cite[Property 2.5(a)]{Chatterjee/Hadi:1988}).
Therefore, the diagonal elements of $H$ are also in $[0,1]$.
Let us check that $h_i$ are in fact in the semi-open interval $[0,1)$ directly,
without using the representation in terms of $\tilde H$.
Representing $K=Q'\Lambda Q$ as above,
where $\Lambda$ is diagonal with nonnegative entries and $Q'Q=I$,
we have
\begin{multline}\label{eq:H-diag}
  H
  =
  (K+aI)^{-1} K
  =
  (Q'\Lambda Q+aI)^{-1} Q'\Lambda Q
  =
  (Q'(\Lambda+aI)Q)^{-1} Q'\Lambda Q\\
  =
  Q^{-1}(\Lambda+aI)^{-1}(Q')^{-1} Q'\Lambda Q
  =
  Q'(\Lambda+aI)^{-1}\Lambda Q.
\end{multline}
The matrix $(\Lambda+aI)^{-1}\Lambda$ is diagonal with the diagonal entries in the semi-open interval $[0,1)$.
Since $Q'Q=I$, the columns of $Q$ are vectors of length 1.
By \eqref{eq:H-diag}, each diagonal element of $H$ is then of the form $\sum_{i=1}^n \lambda_i q_i^2$,
where all $\lambda_i\in[0,1)$ and $\sum_{i=1}^n q_i^2 = 1$.
We can see that each diagonal element of $H$ is in $[0,1)$.

The equality~\eqref{eq:relation} itself was used only for motivation, so we do not prove it;
for a proof in the non-kernelized case, see, e.g., \cite[(4.11) and Appendix C.7]{Montgomery/etal:2012}.

\subsection*{Proof of Lemma~\ref{lem:B}}

In our proof of $B_i>0$ we will assume $a>0$, as usual.
We will apply the results discussed so far in this appendix to the matrix $\bar H$ in place of $H$
and to $n+1$ in place of $n$.

Our goal is to check the strict inequality
\begin{equation}\label{eq:to-check-0}
  \sqrt{1-\bar h_{n+1}}
  +
  \frac{\bar h_{i,n+1}}{\sqrt{1-\bar h_{i}}}
  >
  0;
\end{equation}
remember that both $\bar h_{n+1}$ and $\bar h_i$ are numbers in the semi-open interval $[0,1)$.
The inequality \eqref{eq:to-check-0} can be rewritten as
\begin{equation}\label{eq:to-check}
  \bar h_{i,n+1}
  >
  -\sqrt{(1-\bar h_{n+1})(1-\bar h_{i})}
\end{equation}
and in the weakened form
\begin{equation}\label{eq:true}
  \bar h_{i,n+1}
  \ge
  -\sqrt{(1-\bar h_{n+1})(1-\bar h_{i})}
\end{equation}
follows from \cite[Property~2.6(b)]{Chatterjee/Hadi:1988}
(which can be applied to $\tilde H$).

Instead of the original hat matrix $\bar H$ we will consider the extended matrix \eqref{eq:H-tilde},
where $\tilde X$ is defined by \eqref{eq:extension} with $\bar X$ in place of $X$.
The elements of $\tilde H$ will be denoted $\tilde h$ with suitable indices,
which will run from $-p+1$ to $n+1$, in order to have the familiar indices for the submatrix $\bar H$.
We will assume that we have an equality in \eqref{eq:to-check} and arrive at a contradiction.
There will still be an equality in \eqref{eq:to-check} if we replace $\bar h$ by $\tilde h$,
since $\tilde H$ contains $\bar H$.
Consider auxiliary ``random residuals'' $E:=(I-\tilde H)\epsilon$,
where $\epsilon$ is a standard Gaussian random vector in $\R^{p+n+1}$;
there are $p+n+1$ random residuals $E_{-p+1},\ldots,E_{n+1}$.
Since the correlation between the random residuals $E_i$ and $E_{n+1}$ is
\[
  \corr(E_i,E_{n+1})
  =
  \frac{-\tilde h_{i,n+1}}{\sqrt{(1-\tilde h_{n+1})(1-\tilde h_{i})}}
\]
(this easily follows from $I-\tilde H$ being a projection matrix and is given in, e.g., \cite[p.~11]{Chatterjee/Hadi:1988}),
\eqref{eq:true} is indeed true.
Since we have an equality in \eqref{eq:to-check} (with $\tilde h$ in place of $\bar h$),
$E_i$ and $E_{n+1}$ are perfectly correlated.
Remember that neither row number $i$ nor row number $n+1$ of the matrix $I-\bar H$ are zero
(since the diagonal elements of $\bar H$ are in the semi-open interval $[0,1)$),
and so neither $E_i$ nor $E_{n+1}$ are zero vectors.
Since $E_i$ and $E_{n+1}$ are perfectly correlated,
the row number $i$ of the matrix $I-\tilde H$ is equal to a positive scalar $c$ times its row number $n+1$.
The projection matrix $I-\tilde H$ then projects $\R^{p+n+1}$ onto a subspace of the hyperplane in $\R^{p+n+1}$
consisting of the points with coordinate number $i$ being $c$ times the coordinate number $n+1$.
The orthogonal complement of this subspace, i.e., the range of $\tilde H$,
will contain the vector $(0,\ldots,0,-1,0,\ldots,0,c)$
($-1$ being its coordinate number $i$).
Therefore, this vector will be in the range of $\tilde X$ (cf.\ \eqref{eq:H-tilde}).
Therefore, this vector will be a linear combination of the columns of the extended matrix \eqref{eq:extension}
(with $\bar X$ in place of $X$),
which is impossible because of the first $p$ rows $\sqrt{a}I_p$ of the extended matrix.

\end{document}